\documentclass[11pt,a4paper]{article}
\usepackage[T1]{fontenc}
\usepackage[utf8]{inputenc}
\usepackage{lmodern}
\usepackage[margin=25mm]{geometry}
\usepackage{microtype,amsmath,amssymb,booktabs,multicol,graphicx,float}
\usepackage[font=small,labelfont=bf]{caption}
\usepackage[round,authoryear]{natbib}
\usepackage{xurl}
\usepackage[colorlinks=true,allcolors=blue]{hyperref}
\title{Formalizing Formal Semantics:\\The Formal Atlas}
\author{Stergios Chatzikyriakidis\textsuperscript{1,2} \qquad Aarne Talman\\[4pt]
\small\textsuperscript{1}Institute for Language and Speech Processing, Athena Research Center\\
\small\textsuperscript{2}University of Turku}
\date{November 14, 2026}
\begin{document}
\maketitle
\vspace{-1.5em}
\begin{abstract}
In this paper we present the Formal Atlas, a system for the study of natural language semantics using proof assistants. Our aim is to formalize semantic accounts, check their predictions and compare the implemented theories. To this end, we connect source records and formal developments with an interactive application, where researchers can inspect proofs and modify supported Rocq files. Montague's PTQ and Gallin provide our main cases, followed by dynamic and type-theoretic comparisons, RSA and distributional semantics. We also discuss what it takes to formalize less explicit analyses. Lastly, we consider the use of AI, both for assistance in developing these resources and for inference systems that can make use of them. This work extends an earlier programme in proof-assistant semantics, with research and teaching as complementary uses.
\end{abstract}

\section{Introduction}
In this paper we propose a way to study meaning across diverse semantic frameworks using proof assistants. Different systems do not build the same semantic objects, e.g. truth conditions, information updates, dependent objects or probability distributions. So what can we make explicit across them? In every system, we spell out definitions, the inferences that semantic accounts give rise to, and how accounts relate. Our goal is cumulative work, though with no single shared ontology.

Hales and his colleagues verified the proof of Kepler's conjecture using proof assistants \citep{Hales2017}. Lean's mathlib enables new research to build directly on previous work through shared definitions and theorems \citep{Mathlib2020}. We intend to apply this approach to semantics as well.

This approach is directly related to prior research such as Ranta's type-theoretical grammar \citep{Ranta1994}, Rocq for natural language inference in 2014, and studies proposing broader semantic experiments using proof assistants \citep{ChatzikyriakidisLuo2014,ChatzikyriakidisLuo2016}.

Related research uses Rocq to address inference problems. Bernardy and Chatzikyriakidis connected GF and Rocq for FraCaS \citep{BernardyChatzikyriakidis2017}. Mineshima and his colleagues mapped CCG derivations to logical formulas and used Rocq to prove inferences \citep{MineshimaEtAl2015,MartinezGomezEtAl2016}. The 2016 proposal already included toy semantic grammars for TTR, the Montagovian Generative Lexicon and neo-Davidsonian semantics, alongside MTT accounts.

The Atlas in this study extends these works through detailed source formalization and comparison. Chatzikyriakidis developed the Atlas, and this study links it to collaborative research on inference and evaluation.

\newpage
\section{System design and the application}
A library can hold useful work and still be hard to use. A theorem leaves open which analysis it targets, why its assumptions were chosen and whether editing a definition changes the meaning. Our Atlas interface links these questions to the files, so a researcher or student can start from a source paper, a comparison or a guided example.

\begin{figure}[H]
\centering
\includegraphics[width=16cm]{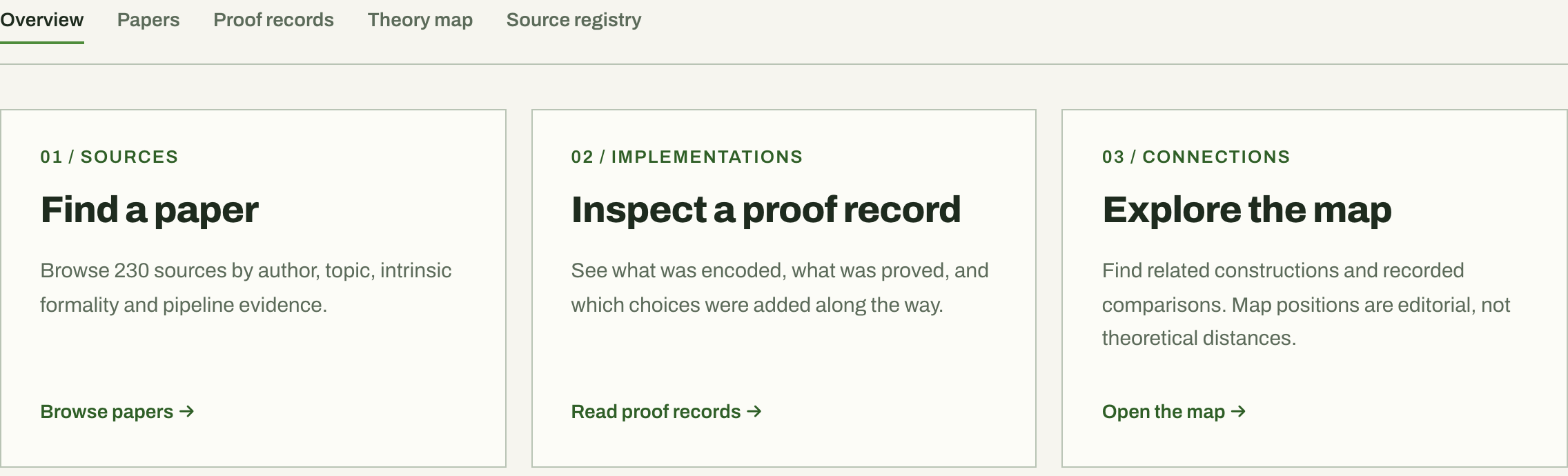}
\caption{The deployed application's entry points: source papers, proof records and the theory map. Screenshot captured on 13 November 2026.}
\label{fig:overview}
\end{figure}

The reader places the claim and assumption records beside the full Rocq source. The declaration index takes the reader to the relevant statement, retaining the definitions around it. There is no need to reconstruct a theorem from its prose description. Its hypotheses and type parameters are there to inspect, and downloading keeps the actual file.

\begin{figure}[H]
\centering
\includegraphics[width=16cm]{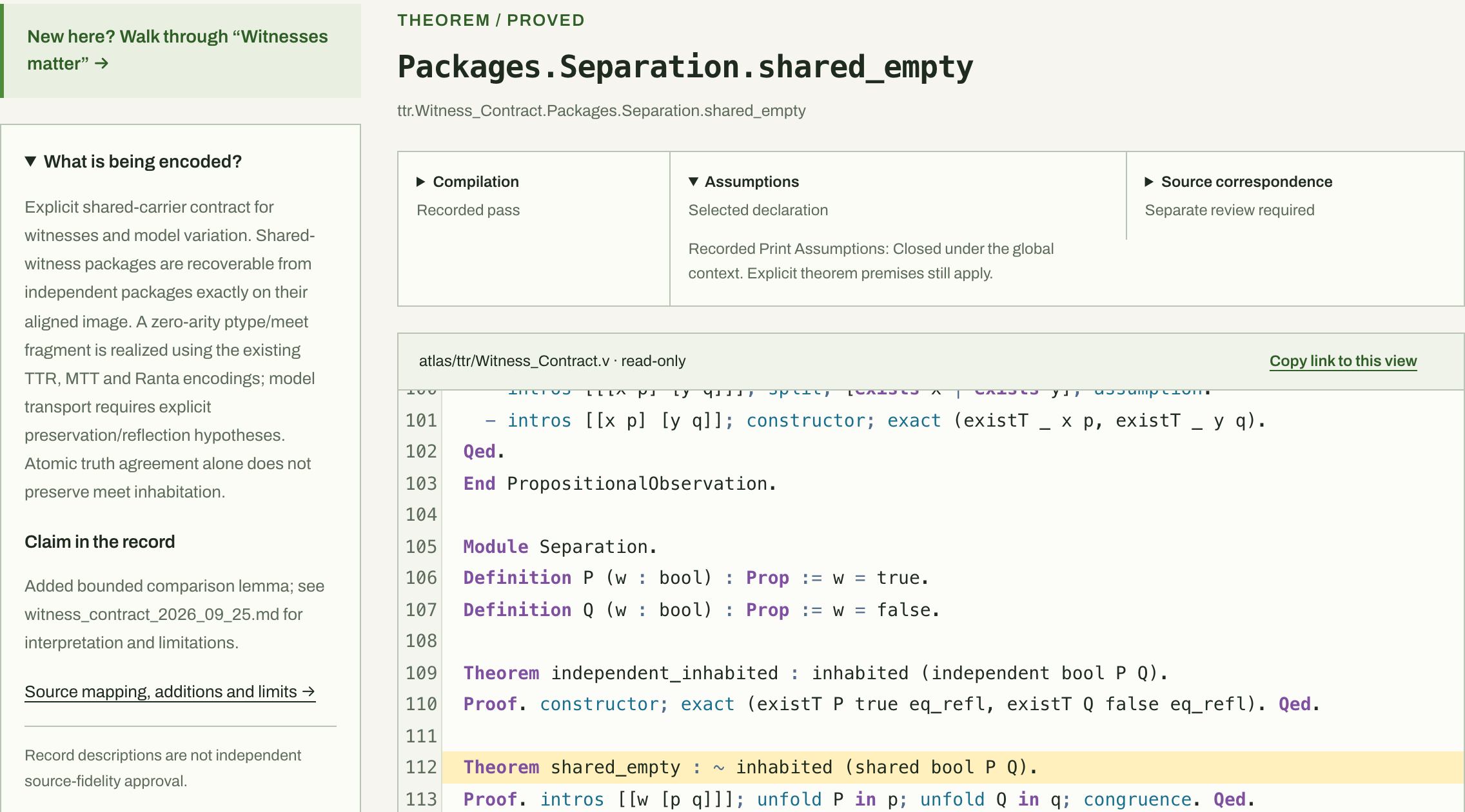}
\caption{The proof reader, with the selected theorem, recorded assumptions, source-correspondence status and the original Rocq code.}
\label{fig:reader}
\end{figure}

\newpage

Our editor then opens a private copy of a supported file. Definitions and proofs can be edited, goals inspected up to a selected command, and the full copy compiled. The imports, however, are precompiled. Editing one file does not rebuild its dependents, so this is not a multi-file IDE. The source-first PTQ and Gallin packages have reading pages and downloads but are not in the public live worker at this revision. For these packages we use local builds.

\begin{figure}[H]
\centering
\includegraphics[width=16cm]{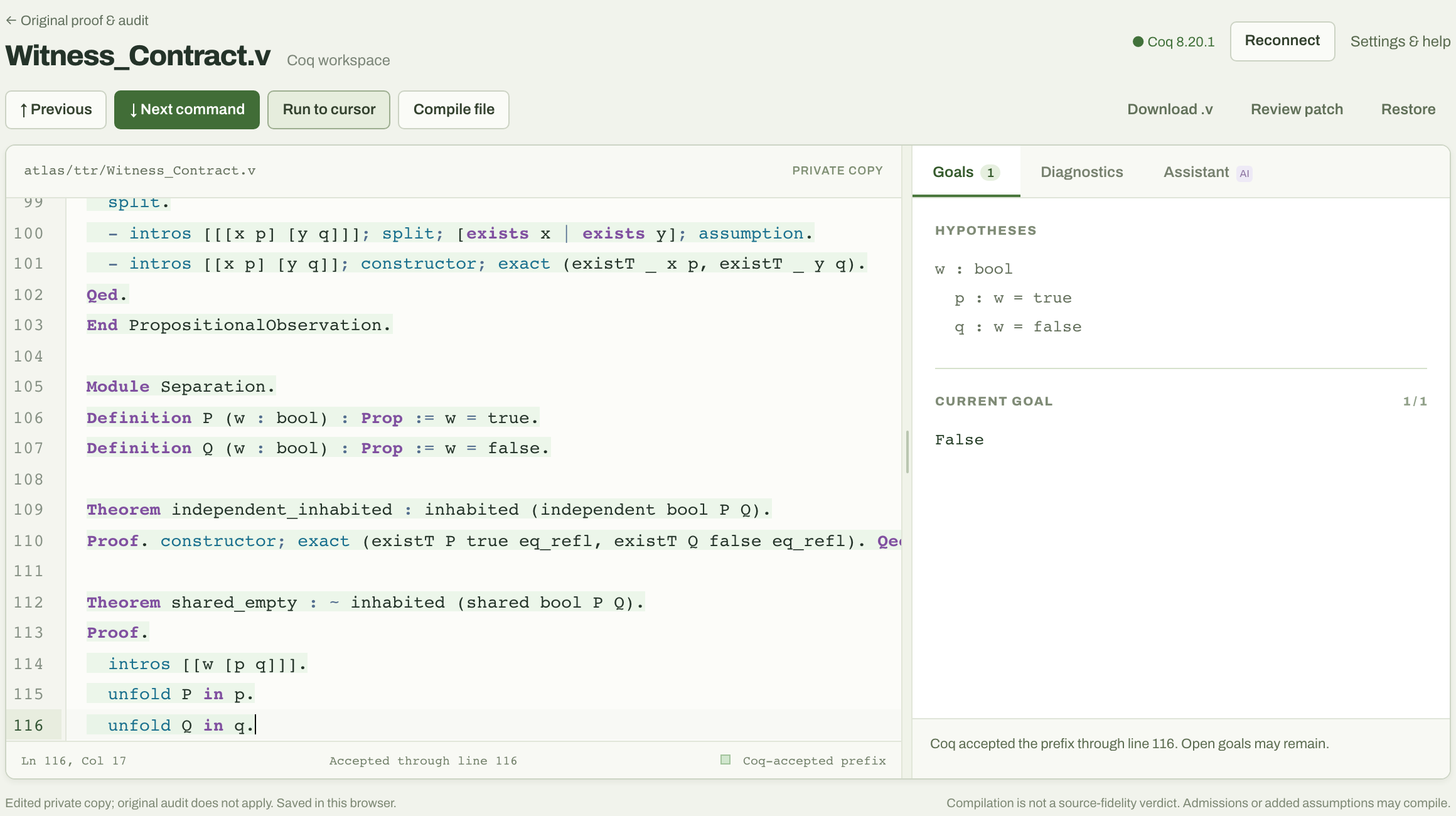}
\caption{The private-copy editor, with proof controls and Goals, Diagnostics and Assistant panels. Connected to Rocq 8.20.1 with an open goal after checking the selected prefix. Captured on 14 November 2026.}
\label{fig:editor}
\end{figure}

The distinction between a proof and a description also applies to AI support. One function offers explanations of selected code in the repo, while a different one is more experimental and takes a claim, supplies a model with the vocabulary of selected modules, and checks the proposed Rocq fragment. The model can use compiler feedback for another attempt. Code and queried assumptions remain available for inspection. Needless to say that a model's outcome label does not show that its theorem expresses the user's claim.

What is recorded in the Atlas itself? At revision \texttt{78404c4}, 230 survey entries are linked to formal descriptions, comparisons and assessments \citep{Atlas2026}. Of these, 43 have Rocq-file links, while there are 87 file-level assessments. A surveyed source is not thereby a completed formalization. We keep the source's explicitness separate from the progress of implementation. Independent human source review remains necessary even for compiled developments.

The visitor needs no model subscription to read files or check Rocq. Optional model calls use the visitor's OpenRouter key. A private edit inherits no source assessment, and a green prefix marker may leave proof goals open. A downloaded file or patch can be proposed for review, but it is not automatically published.

For teaching, we can use the same material at different levels. Students can work out a prediction on paper before inspecting the file. They can then change an assumption and look for a proof or countermodel. Research students can continue the development in that file. In each case, successful checking must be accompanied by an explanation of the semantic result.

\newpage
\section{Montague and Gallin: keeping the formal layers explicit}
In PTQ, grammar rules produce the English analyses, translation gives their IL terms, and model conditions restrict the interpretations \citep{Montague1973}. A formula can be correctly interpreted without checking either its connection to the grammar or those restrictions. We therefore keep grammar, IL, translation and admissibility as separate parts.

The package implements S1--S17 and T1--T17. One interpretation goes through translation into IL,  another  one operates on the analysis tree itself. We prove that the two independently defined operations agree. Thus there is a theorem connecting the layers, rather than an informal decision to regard two computations as equivalent.

Consider now the famous examples  \emph{the temperature is ninety} and \emph{the temperature rises}. Do these two sentences  entail \emph{ninety rises}? The relevant distinction is between agreement at the present index and identity of concepts. In our one-world model the temperature concept is $a$ at one time and $b$ at the next. The concept associated with \emph{ninety} is constantly $a$. A predicate of concepts can distinguish them, even though their current values agree.

The countermodel theorem starts from the English trees, proves the translated premises true and the conclusion false, and establishes all nine implemented families of meaning postulates. This is what makes it a countermodel for the permitted PTQ interpretations. We are not proposing a physical account of rising temperature.

Our initial encoding of Montague got two source clauses wrong. An extra intension in the IV result type forced extension operators into sentences, and necessity quantified only over worlds at the current time. The proofs went through in both cases. The error became apparent when we checked against PTQ. The later package proves that the literal surface rules cannot generate two printed examples and records a separate amendment. It also supplies an additional reading of \emph{Mary believes that John finds a unicorn and he eats it}, with conjunction outside the belief clause. Exhaustive reading classification remains open.

Gallin's comparison needs another separation \citep{Gallin1975}. Making indices explicit in Ty2 gives a translation of IL terms: intension introduces an index argument, extension supplies it, and necessity quantifies over it. We prove interpretation preservation under corresponding models and assignments. But a semantic theorem alone gives no translation of proofs.

Our forward compiler takes proofs in the implemented untensed IL calculus, including assumptions, to proofs in Ty2. It is a compiler of object-language derivations inside Rocq. For the separate tensed comparison, the future relation must link the same world to a strictly later time, and a countermodel shows the effect of dropping that condition. Recorded Rocq 8.20.1 builds and kernel rechecks cover 443 PTQ and 225 Gallin statements, with assumption queries. No reverse proof compiler is claimed.

\newpage
\section{What Is Preserved and What Is Lost if We Compare}
When making comparisons, one must first clarify what `agreement' means. Is it agreement w.r.t.\ truth, discourse referents, or structured objects inhabiting a specific type?

These questions are different, and the Atlas contains examples where overlooking this distinction leads to incorrect results.

In file change semantics \citep{Heim1982}, a file contains both world information and assignments to discourse referents. Consider the sentence \emph{The king did not eat lunch}. Local accommodation introduces the king within the negated update, whereas global accommodation introduces the king into the external context. In both cases, a true file can be left in the same model. However, introducing the king into the outer context allows subsequent pronouns to use the king as a referent, whereas introducing him within the negated update does not. With no king, the policies can also differ in truth. Therefore, current truth conditions alone cannot resolve the comparison.

Our DPL development provides counterexamples to claims as printed in \citet{GroenendijkStokhof1991}. For example, unrestricted disjunction idempotency does not hold for existential updates. $\varphi\lor\varphi$ tests whether $\varphi$ can succeed but does not retain its changed assignment. The law restricted to tests is proved.

Similarly, our TTR--MTT case is not about one framework retaining witnesses while the other lacks them, but rather about a specific map \citep{Cooper2023,ChatzikyriakidisLuo2020}. In a fixed model, we link the existence of a TTR record containing an individual and evidence that he is a man and runs to an existential proposition in the relevant MTT interpretation.

Assume that only the individual is projected. Since two records with different evidence can have the same individual, it is not possible to recover both originals by reversing the projection. Here, nothing prevents MTT from retaining evidence in dependent packages. The loss occurs in our map. The intersection example raises a related question: distinct objects inhabiting two types do not necessarily provide a single object inhabiting both types. The requirement that the object is shared must be maintained.

In DTS, common nouns are predicates on entities \citep{BekkiMineshima2017}. From $N$, we form $\sum_{x:E}N(x)$, which pairs an entity with evidence of noun membership. Starting instead from a noun type, we select a map into an entity domain and take its image predicate. The existential and universal results state exactly the correspondences obtained from these choices.

Then, our small resolver searches for entities provided by the typed discourse context. When looking for an entity, it skips evidence fields but does not remove that evidence from the context. Two objects remain as possible reference targets. Selecting the first object is an implemented policy, not an explanation of salience. The failure of this search implies that there are no candidates within the limited search space and should not be interpreted as a general failure of DTS.

A comparison to Kratzer's preliminary universal clause can be easily obtained by defining accessibility as the satisfaction of all modal-base propositions \citep{Kratzer1977}. Expanding the definitions yields the same clause. This confirms a choice of representation, but in itself tells us nothing about theoretical equivalence, ordering sources, inconsistent backgrounds, or tense.

\newpage
\section{Other Semantics and Less Explicit Analysis}
Semantic results do not necessarily have to be truth values. Our probabilistic and distributional modules are designed to test the same programme using different mathematical objects. Therefore, what needs to be verified depends on the desired result. Examples include probability distributions or sentence vectors.

Bayesian research by Bernardy, Blanck, Chatzikyriakidis, Lappin, and Maskharashvili includes compositional semantics, modeling systems, inference tests, and predicate representations \citep{BernardyEtAl2018,BernardyEtAl2019,BernardyPredicates2019}. Grove, Bernardy, and Chatzikyriakidis connect compositional semantics and Bayesian pragmatics through logical inference \citep{GroveEtAl2021}. The finite RSA development introduced in this paper should not be confused with mechanizing all of this research.

RSA provides a recursive relationship between listener and speaker \citep{GoodmanFrank2016,BergenEtAl2016}. The literal listener uses truth conditions, the speaker considers alternative utterances, and the pragmatic listener reasons about the speaker. These calculations can be verified without sampling using exact rational arithmetic. For zero, one, and two objects, with alternatives \emph{none}, \emph{some}, and \emph{all}, literal \emph{some} gives equal probability to the latter two worlds. With a uniform prior distribution, equal utterance weights, and $\alpha=1$, the pragmatic probabilities are $3/4$ and $1/4$.

We prove these values and show the conditions under which normalization yields a probability distribution. This allows us to separate changes in prior distribution and interpretation from numerical errors when estimating the same quantity. Further connections to Bayesian semantics are a subject for future research.

In DisCoCat \citep{CoeckeEtAl2010}, grammatical reduction determines the tensor operations that generate the sentence vector. Since the subject and object occupy different argument positions, swapping them can change the vector. Our fragment proves that the calculated meanings match for equivalent reductions. This does not determine lexical vectors or establish entailment from similarity. Furthermore, the current general-purpose interface does not enforce every connection between a lexical type sequence and its reduction.

For Derrida (1972, English translation \citeyear{Derrida1988}), we instead develop an explicitly source-inspired account of forms, occurrences and contexts. In Derridean fashion, there is a multitude of interpretations of the essay, and this formalization is one of them. Its counterexample shows that infinitely many quotation contexts can keep the same interpreted effect, so the number of contexts alone does not give a change in meaning. This concerns our added quotation syntax, not Derrida's non-saturation thesis, which we do not formalize. The idea is that one should be able to formalize and check useful ideas and check predictions or use them in connection to other theories, even if these ideas have been proposed in research traditions that are sceptical of formalization.

\newpage
\section{AI, Related Research, and Future Developments}
Large language models have assisted us throughout the Atlas, helping us with the formulation and formalisation of definitions, declaration search, proof construction, and repair. In particular we have experimented with models GPT-6 Astra from OpenAI and Claude Fable 5.1 and 5.2 using Codex and Claude Code respectively (and various effort levels). Building a resource of this scope unaided would have been difficult for a single developer. We have not quantitatively measured the time saved. While the use of proof assistants for natural language semantics began many years ago, LLM assistance has expanded the scope of what one can actually attempt both in scope and in scale.

The connection to AI also extends in the reverse direction. In our research on generalization and data corruption in natural language inference (NLI), we examine whether benchmark success reflects the inference learned by the system \citep{TalmanChatzikyriakidis2019,TalmanEtAl2021}. By using formal resources, we can set up controlled problems where proofs or countermodels are available. We propose comparing model-only, retrieval-assisted, and proof-supported systems, using changes that preserve inference and changes that reverse it.

Such tests require measuring translation and proof search separately. A verified proof of an incorrect translation still only answers the wrong question. Failure to find a proof establishes neither falsehood nor neutrality. Similarly, a countermodel to entailment does not prove entailment of the opposite conclusion. Coverage and cost are important alongside correct answers. Research on uncertainty-aware NLI raises further questions about human interpretations \citep{TalmanEtAl2023}. These cannot simply be equated with neural confidence or semantic probability.

Linglib develops a related cumulative programme in Lean to integrate analyses from various fields of linguistics into a shared library containing definitions and machine-checked results \citep{Linglib2026}.

The emphasis on reusable formal resources overlaps with a key objective of our programme. A comparison of the two research programmes is something that is of great interest to us and as such we plan it for future work.

Immediate work involves independent source reviews, more detailed comparisons, Bayesian connections, portability, and evaluation, as well as more coverage. The \href{https://github.com/StergiosCha/formal_atlas_of_semantics/tree/78404c4c9ab266cb07c0539f7ff39f47c669f1db}{repository} and \href{https://salmon-meadow-0d05c8710.5.azurestaticapps.net/}{web application} make the materials available to researchers and teachers. The results presented here pertain to \texttt{78404c4}, and the live service is subject to change. We will update the results presented here accordingly. Our aim is to provide resources that others can inspect and extend.

\smallskip
{\footnotesize\textit{Acknowledgements.} The first author gratefully acknowledges support from the Human Diversity consortium, under the Profi7 program (grant no.~352727) and the Centre of Excellence (grant no.~374221) by the Research Council of Finland.}

\smallskip
{\footnotesize\textit{AI use.} OpenAI Codex assisted with this shortened report and code inspection. Previous development and assessment also used Claude and Gemini. The authors are responsible for the final text and claims.}

\clearpage
\renewcommand{\bibsection}{\section*{References}}
\setlength{\bibsep}{3pt}
\begin{multicols}{2}
\footnotesize
\raggedright
\sloppy
\bibliographystyle{plainnat}
\bibliography{\jobname}
\end{multicols}
\end{document}